\documentclass[11pt]{article}
\usepackage[final]{acl} 

\usepackage[T1]{fontenc}
\usepackage[utf8]{inputenc}
\pdfoutput=1
\usepackage{times}
\usepackage{latexsym}
\usepackage{microtype}
\usepackage{inconsolata}
\usepackage{graphicx}
\usepackage{amsmath}
\usepackage{amssymb}
\usepackage{booktabs}
\usepackage{enumitem}
\usepackage{hyperref} 
\usepackage{subcaption} 

\usepackage{float}
\usepackage{makecell}
\usepackage{array}
\usepackage[english]{babel}
\usepackage{inconsolata}

\title{WixQA: A Multi-Dataset Benchmark for Enterprise Retrieval-Augmented Generation}

\author{
Dvir~Cohen \\
\texttt{dvirco@wix.com}
\And
Lin~Burg \\
\texttt{linb@wix.com}
\And
Sviatoslav~Pykhnivskyi \\
\texttt{sviatoslavp@wix.com}
\And
Hagit~Gur \\
\texttt{hagitg@wix.com}
\AND
Stanislav~Kovynov \\
\texttt{stanislavk@wix.com}
\And
Olga~Atzmon \\
\texttt{olgaa@wix.com}
\And
Gilad~Barkan \\
\texttt{giladba@wix.com} \AND
Wix.com AI Research, Tel Aviv, Israel
}

\date{} 

\begin{document}

\maketitle

\begin{abstract}
Retrieval-Augmented Generation (RAG) is a cornerstone of modern question answering (QA) systems, enabling grounded answers based on external knowledge. Although recent progress has been driven by open-domain datasets, enterprise QA systems need datasets that mirror the concrete, domain-specific issues users raise in day-to-day support scenarios. Critically, evaluating end-to-end RAG systems requires benchmarks comprising not only question--answer pairs but also the specific knowledge base (KB) snapshot from which answers were derived. To address this need, we introduce \textbf{WixQA}, a benchmark suite featuring QA datasets precisely grounded in the released KB corpus, enabling holistic evaluation of retrieval and generation components. WixQA includes three distinct QA datasets derived from Wix.com customer support interactions and grounded in a snapshot of the public Wix Help Center KB: (i) \textit{WixQA-ExpertWritten}, 200 real user queries with expert-authored, multi-step answers; (ii) \textit{WixQA-Simulated}, 200 expert-validated QA pairs distilled from user dialogues; and (iii) \textit{WixQA-Synthetic}, 6,221 LLM-generated QA pairs, with one pair systematically derived from each article in the knowledge base. We release the KB snapshot alongside the datasets under MIT license and provide comprehensive baseline results, forming a unique benchmark for evaluating enterprise RAG systems in realistic enterprise environments.

\end{abstract}

\section{Introduction}
Large-scale open-domain question answering (QA) datasets such as SQuAD \cite{squad}, Natural Questions \cite{nq}, TriviaQA \cite{triviaqa}, and InsuranceQA \cite{feng2015insuranceqa} have driven tremendous progress in AI systems based on neural networks designed for question answering. Recently, QA systems are increasingly adopting a Retrieval-Augmented Generation (RAG) framework, in which a retriever first identifies the most relevant documents and then a generator uses the retrieved text to produce candidate answers \cite{lewis2020rag}. However, many real-world settings---especially in enterprise customer support---demand domain-specific evaluation, robust retrieval from curated knowledge bases, and step-by-step guidance rather than a one-shot answer. Central to these real-world scenarios is the need to synthesize information from multiple documents to fully address user queries, a capability that our datasets are designed to evaluate.

Enterprise queries present unique challenges, often demanding complex procedural guidance and specialized vocabulary (characteristics of Long-form Question Answering, LFQA), rather than the short, factual answers targeted by many existing QA benchmarks \cite{squad,nq,triviaqa}. To address this specific need, we introduce \textbf{WixQA}, a benchmark suite designed for the enterprise domain. Recognizing the limitations of existing resources, WixQA adopts a two-pronged design. This approach explicitly incorporates datasets featuring long-form, multi-step answers crucial for troubleshooting and task resolution in enterprise contexts, alongside datasets with shorter, specific answers reflecting more straightforward real-world scenarios. Such diversity makes our benchmark especially versatile for developing robust QA models adept at handling a wide spectrum of response types.

WixQA realizes this vision with three complementary datasets derived from Wix.com support interactions:
\begin{itemize}[noitemsep,topsep=2pt]
    \item \textbf{WixQA-ExpertWritten:} A collection of 200 genuine customer queries paired with step-by-step answers manually authored and validated by domain experts, reflecting real-world support challenges. 
    \item \textbf{WixQA-Simulated:} A set of 200 examples distilled from multi-turn user–chatbot dialogues into clear, single-turn QA pairs, each meticulously validated for procedural accuracy by domain experts through simulation. 
    \item \textbf{WixQA-Synthetic:} Comprising 6,221 question–answer pairs automatically extracted using Large Language Models (LLMs) from Wix articles, this dataset offers scale and diversity for training robust retrieval models.
\end{itemize}

Recent works such as TechQA \cite{techqa}, FinTextQA \cite{fintextqa}, FINQA \cite{finqa}, PubMedQA \cite{jin2019pubmedqa}, WikiHowQA \cite{bolotova2023wikihowqa}, and AmazonQA \cite{gupta2019amazonqa} underscore the trend toward benchmarks that capture expert curation, automatic extraction, and domain-specific nuances. Their diverse scales and focuses further motivate our enterprise-scale approach.

\noindent Our key contributions are as follows:
\begin{itemize}[noitemsep, topsep=2pt]
    \item \textbf{A Diverse Suite of Enterprise QA Datasets:} Introduction of \textbf{WixQA}, comprising three complementary, KB-grounded datasets (ExpertWritten, Simulated, Synthetic) reflecting realistic enterprise support interactions and varying curation methods.
    \item \textbf{Multi-article dependency:} A unique aspect of our datasets is that both ExpertWritten and Simulated datasets' answers can be based on more than one article within the Wix knowledge base. This adds complexity to the RAG process, as the model must retrieve and synthesize information from multiple sources to generate a comprehensive and accurate response. This multi-article dependency reflects real-world scenarios where user queries require integration of knowledge from various parts of the corpus.
    \item \textbf{Enterprise-Scale Knowledge Base:} We release a unified corpus of 6,221 Wix help articles, forming a domain-specific knowledge base that reflects the nuanced language and multi-step procedures typical of enterprise support workflows.
    \item \textbf{Comprehensive Benchmarking:} Extensive experiments across all three datasets provide in-depth analyses of retrieval and generation performance in realistic enterprise environments, highlighting the challenges and opportunities for further advancements.
\end{itemize}

In the following sections, we review related work (\S\ref{sec:related_work}), detail our data collection and annotation protocols in addition to comprehensive statistics for each dataset (\S\ref{sec:data_collection}), and showcase baseline experiments (\S\ref{sec:baseline_experiments}). We conclude with a discussion of findings and directions for future work (\S\ref{sec:conclusion}).

\section{Related Work}
\label{sec:related_work}

\begin{table*}[t]
\centering
\renewcommand{\arraystretch}{1.3}

\setlength{\tabcolsep}{1.2pt}
\begin{tabular}{l c c c c c}
\toprule
\textbf{Dataset} & \textbf{Domain} & \textbf{Questions Source} & \textbf{Answers} & \textbf{Knowledge Base} & \textbf{Size} \\
\midrule
TechQA & Tech Support & Expert-selection & Expert-written & IBM Technotes & 600 \\
BioASQ-QA & Biomedical & Manual & Expert-written & PubMed & 500 \\
FinTextQA & Finance & Auto-generated & Auto-generated & Finance Textbooks & 1,262 \\
FinQA & Finance & Expert-curated & Expert-written & Financial Docs & 8,200 \\
EmrQA & Clinical & Auto-extracted & Auto-extracted & EMRs & 2M \\
DomainRAG & Education & Hybrid & Hybrid & University Pages & 395 \\
Doc2Dial & Gov. Services & Dialogues & Expert-written & Domain Docs & 4,500 \\
WikiHowQA & Procedural & Expert-selected & Expert-written & WikiHow Articles & 11,746 \\
DoQA & Community FAQs & Human-curated & Human-curated & Stack Exchange & 10,917 \\
ConvRAG & Conversational & Hybrid & Expert-validated & Conversations & 1,000 \\
\midrule
WixQA-EW & Enterprise  & User queries & Expert-written & Wix Articles & 200 \\
WixQA-Simulated & Enterprise & User queries & Expert-validated & Wix Articles & 200 \\
WixQA-Synthetic & Enterprise & Auto-generated & Auto-generated & Wix Articles & 6,221 \\
\bottomrule
\end{tabular}
\caption{
\label{tab:related_benchmarks}
Extended Overview of QA Benchmarks including WixQA; for datasets with train–test splits, the reported size refers to the test set}
\end{table*}

Recent advancements in question answering (QA) have seen a significant shift from general-purpose, Wikipedia-based benchmarks toward specialized, domain-specific datasets. This shift is motivated by the increasing demand for practical applications in enterprise environments, where retrieval accuracy, domain-specific vocabulary, and procedural or multi-step reasoning play critical roles.While earlier benchmarks (e.g., SQuAD~\cite{squad}, TriviaQA~\cite{triviaqa}, Natural Questions~\cite{nq}) centered on open-domain factual knowledge from Wikipedia, real-world QA increasingly requires domain-adapted datasets.

Several notable QA benchmarks have emerged addressing needs in organizational and enterprise contexts, focusing on customer support, internal documentation, and procedural guidance. TechQA~\cite{techqa} is particularly relevant, comprising real user-generated questions from IBM's technical support forums paired with answers located within IBM's corpus of technical documents (Technotes). It provides both short and long-form answers, making it suitable for evaluating retrieval accuracy within an enterprise technology support context. Similarly, Doc2Dial~\cite{doc2dial} and MultiDoc2Dial~\cite{multidoc2dial} focus on customer support dialogues grounded in domain-specific documents (sourced from public sector entities such as government agencies), evaluating conversational RAG systems that track information over multiple turns.
 These benchmarks highlight the importance of using authentic \emph{organizational data and domain relevance}, a principle shared by our WixQA datasets, which are exclusively derived from Wix customer support interactions and its carefully curated repository of Wix articles. This ensures evaluation on realistic enterprise support scenarios involving multi-step procedures and domain-specific terminology.

Procedural knowledge is another key focus area. WikiHowQA~\cite{bolotova2023wikihowqa} leverages WikiHow articles to evaluate QA systems on step-by-step instructions. Although not strictly organizational data, its procedural nature aligns closely with enterprise troubleshooting scenarios, emphasizing detailed, actionable answers. Datasets like DoQA~\cite{doqa} have introduced FAQ-style conversational QA across multiple Stack Exchange domains, reflecting realistic interactions in specialized communities and the complexity of retrieving multi-step answers.

The quality and curation method of answers vary across benchmarks. Expert curation is a hallmark of quality in specialized QA benchmarks like TechQA and FinQA~\cite{finqa}, which rely heavily on expert-validated answers. Doc2Dial and MultiDoc2Dial provide annotated dialogues grounded in organizational documents. Synthetic datasets, such as FinTextQA~\cite{fintextqa} (focused on finance), leverage automatic methods for large-scale coverage. WixQA adopts a multi-faceted approach: WixQA-ExpertWritten and WixQA-Simulated feature rigorous expert validation to ensure procedural accuracy, similar to TechQA and Doc2Dial. Meanwhile, WixQA-Synthetic, auto-generated from expert-written Wix articles, offers extensive training data while maintaining high preliminary accuracy, balancing scale with precision akin to synthetic benchmarks. Further expanding the evaluation landscape, datasets like DomainRAG~\cite{domainrag} and ConvRAG~\cite{convrag} incorporate hybrid curation methods and address aspects like multilingual and conversational RAG, highlighting the growing diversity in benchmark design

The supporting knowledge base characteristics significantly impact RAG performance. Domain-focused benchmarks such as Doc2Dial, MultiDoc2Dial, and FinQA highlight how curated organizational documents (like customer support documents or financial statements) enhance retrieval precision and answer relevance. WixQA similarly provides a unified corpus of 6,221 Wix articles, forming a cohesive, domain-specific knowledge base. This approach rigorously evaluates models' ability to retrieve highly relevant procedural content from organizational documentation, a crucial capability for enterprise RAG systems.

The WixQA datasets introduced in this work uniquely contribute to this landscape by blending expert-written (ExpertWritten), expert-validated distilled (Simulated), and LLM-generated (Synthetic) QA pairs, all derived from Wix’s internal knowledge base. A key distinction is the explicit inclusion of multi-article dependencies in the ExpertWritten and Simulated datasets, requiring synthesis across documents—a common real-world enterprise challenge. Table~\ref{tab:related_benchmarks} overviews these benchmarks alongside WixQA, highlighting their diverse characteristics.

\section{Data Collection and Annotation}
\label{sec:data_collection}
This section describes the methodology for creating our three datasets and our knowledge base. Each dataset was designed to address different aspects of enterprise QA. The creation of answers, particularly those involving expert curation, adhered to specific instructions (see Appendix~\ref{app:expert_authoring_guidelines}) aimed at ensuring responses were comprehensive, KB-grounded, and user-centric.

\subsection{Knowledge Base}
\label{sec:knowledge_base}

The Knowledge Base (KB) supporting our benchmark contains a snapshot of English-only articles from Wix's customer support repository\footnote{https://support.wix.com/en}. The KB consists of 6,221 articles distributed across three distinct types:

\begin{itemize}[noitemsep,topsep=2pt]
    \item \textbf{Article (66\%):}  General-purpose articles covering a wide range of topics, including tutorials, troubleshooting guides, and feature explanations.
    \item \textbf{Feature Request (32\%):} Articles that provide information about unsupported features, allow users to vote for them, and announce when a feature has been implemented.
    \item \textbf{Known Issue (1\%):} Articles that document known problems acknowledged by Wix and provide updates on their resolution status.
\end{itemize}

\begin{table*}[ht]
\centering
\label{tab:dataset_stats_summary}
\begin{tabular}{l r c c c}
\toprule
Dataset & Size & Question Tokens & Answer Tokens & Multi-Article \% \\
\midrule
ExpertWritten & 200 & 19 & 172 & 27\% \\
Simulated & 200 & 12 & 50 & 14\%\\ 
Synthetic & 6,221 & 24 & 130 & 0\% (by design) \\
\bottomrule
\end{tabular}
\caption{\label{tab:ds_stats}
Key Statistics for WixQA Datasets (Median Values)}

\end{table*}

The Knowledge Base (KB) is a crucial component of RAG applications, providing the necessary context to ground the LLM's responses. This ensures answers rely on specific, up-to-date enterprise information, particularly where the LLM's internal knowledge may be absent, outdated, or limited.

Table~\ref{tab:ds_stats} summarizes key statistics for the three WixQA datasets, highlighting differences in scale, question/answer complexity, and multi-article dependency.

The higher multi-article percentage in ExpertWritten (27\%) compared to Simulated (14\%) reflects the former's design for comprehensive expert solutions versus the latter's focus on concise, distilled answers.

\subsection{WixQA-ExpertWritten - Real User Queries with Expert Answers}
\textit{WixQA-ExpertWritten}, the first dataset in our suite, contains 200 authentic customer queries from Wix.com support interactions. Each query is paired with a detailed, step-by-step answer meticulously authored by domain experts. An example of a question and an answer is provided in Appendix~\ref{app:ExpertWritten_example}.
Key characteristics and the creation process include:

\begin{itemize}[noitemsep,topsep=2pt]
    \item \textbf{Source of Questions:} Genuine user queries submitted through Wix support channels, covering a diverse range of real-world issues such as domain configuration, SSL certificate troubleshooting, editor functionalities, etc.
    \item \textbf{Expert Answer Curation:} Ground truth answers were manually authored by Wix support experts. These answers are designed to be comprehensive and granular, often providing detailed step-by-step instructions necessary for resolving complex user problems and ensuring users have a clear, actionable solution. This authoring process followed the specific instructions outlined in Appendix~\ref{app:expert_authoring_guidelines}, which emphasize fidelity to KB content, completeness, and user-focused communication.
    \item \textbf{Knowledge Base Grounding and Multi-Article Synthesis:} Answers are grounded in the official Wix knowledge base (\S\ref{sec:knowledge_base}). Crucially, many answers require synthesizing information from multiple KB articles (27\% involve more than one article) to fully address the user's query, reflecting realistic enterprise support scenarios where information is often distributed across documents.
    \item \textbf{Rigorous Manual Validation:} A multi-stage validation process ensured answer quality and accuracy. Initially drafted answers were reviewed by three peer experts; acceptance required a majority vote. Subsequently, two senior experts conducted a final validation across the entire dataset. Answers failing to meet strict criteria for factual accuracy, clarity, and relevance to the query were removed at either stage. This rigorous protocol ensures the dataset reflects production-level support standards.
    \item \textbf{Intended Use Case:} This dataset is ideal for evaluating the ability of RAG systems to handle authentic, potentially complex user queries that necessitate detailed, multi-step, and sometimes multi-source answers. It tests the system's capacity for both accurate retrieval (potentially across multiple documents) and comprehensive generation. 
    \item \textbf{Statistics:} As summarized in Table~\ref{tab:ds_stats}, this dataset features concise real user queries (median 19 tokens) paired with comprehensive, expert-authored answers (median 172 tokens). This difference highlights the need for detailed, step-by-step guidance. The significant context size (median 5,928 tokens per answer's source articles) and multi-article ratio (27\% require >1 article) underscore the complexity addressed.

\end{itemize}
\subsection{WixQA-Simulated: Expert-Validated QA Pairs from User Conversations}
Complementing the \textit{WixQA-ExpertWritten} dataset, \textit{WixQA-Simulated} offers 200 QA pairs between Wix users and support chatbot dialogues, refined for conciseness and procedural accuracy via expert validation. It targets evaluating RAG systems on generating comprehensive yet effective guidance. An example is provided in Appendix~\ref{app:Simulated_example}.

The creation process involved several key stages:
\begin{itemize}[noitemsep,topsep=2pt]
    \item \textbf{Source Material:} We collected multi-turn conversational dialogues between Wix users and the support chatbot addressing specific issues.
    \item \textbf{QA Pair Distillation:} Using an LLM, these dialogues were distilled into single-turn question--answer pairs. The objective was to capture the core user problem and the essential steps of the expert's solution concisely.
    \item \textbf{Knowledge Base Grounding:} Answers were grounded in the Wix knowledge base (described in \S\ref{sec:knowledge_base}). Similar to \textit{WixQA-ExpertWritten}, answers may require synthesizing information from multiple KB articles, but the emphasis here is on brevity and directness in the final distilled answer.
    \item \textbf{Rigorous Multi-Stage Validation:} 
    The distilled QA pairs underwent a stringent validation process. First, automatic filtering removed irrelevant QA (e.g., requests for human agents, ambiguous queries, questions unanswerable solely by the KB). Second, three domain experts manually reviewed the remaining pairs, discarding those where the answer did not appropriately address the question. Finally, for the pairs passing these filters, annotators performed simulation-based validation: they meticulously followed the step-by-step instructions in each answer to verify its correctness and confirm that it resolved the user's specific problem. This multi-stage process ensured the final set of 200 QA pairs is both highly relevant and procedurally accurate.
    \item \textbf{Intended Use Case:} This dataset serves as a benchmark for evaluating a RAG system's ability to generate specific, concise, and accurate responses, particularly for procedural or multi-step tasks. It contrasts with \textit{WixQA-ExpertWritten}'s focus on comprehensive answers, instead testing for maximal accuracy within minimal length.
    \item \textbf{Statistics:} Designed for conciseness, this dataset features short questions (median 12 tokens) and relatively brief answers (median 50 tokens), as shown in Table~\ref{tab:ds_stats}. This contrasts with the ExpertWritten dataset and reflects the goal of evaluating accurate, distilled guidance. A notable portion (14\%) still requires multi-article synthesis. 
\end{itemize}

\subsection{WixQA-Synthetic: Large-Scale QA Pairs via LLM-Based Extraction}
To complement the expert-curated datasets and provide large-scale data suitable for training robust models, we created \textit{WixQA-Synthetic}. This dataset comprises 6,221 question--answer pairs, with one pair generated for each of the 6,221 articles in the Wix.com knowledge base.An example is provided in Appendix~\ref{app:Synthetic_example}.

The generation and validation process involved the following steps:
\begin{itemize}[noitemsep,topsep=2pt]
    \item \textbf{Automated QA Generation:} We applied a state-of-the-art LLM, specifically GPT-4o, to each of the 6,221 articles within our knowledge base (\S\ref{sec:knowledge_base}). Recognizing the distinct structure and purpose of the three article types (Article, Feature Request, Known Issue), we employed tailored prompts for each type to optimize the quality of the extracted QA pairs. An example prompt for the 'Feature Request' type is provided in Appendix~\ref{app:qa_extraction_feature_request}.
    \item \textbf{Single-Article Grounding:} By design, each generated question--answer pair is explicitly linked to the single source knowledge base article from which it was derived. This provides a direct ground truth reference (article URL) for evaluating retrieval performance using common metrics like Precision@K and Recall@K. This single-source grounding contrasts with the potential multi-article synthesis required for the \textit{WixQA-ExpertWritten} and \textit{WixQA-Simulated} datasets.
    \item \textbf{Quality Assurance and Validation:} To assess the quality of the LLM-generated data, two domain experts manually evaluated a random sample of 250 QA pairs. This review confirmed high fidelity, with over 90\% of the sampled answers found to be correct and relevant to the source article. Building on this positive result, we performed manual sanity checks across the entire generated set (6,221 pairs) to ensure overall data integrity and absence of major formatting errors or inconsistencies. This multi-faceted validation confirmed the reliability of the extraction pipeline.
    \item \textbf{Intended Use Case:} With its substantial size, this dataset is particularly well-suited for training or fine-tuning the retrieval and generation components of RAG systems. It offers broad coverage of the knowledge base content, enabling models to learn domain-specific patterns and terminology at scale.
    \item \textbf{Statistics:} With 6,221 QA pairs, this dataset offers considerable volume for training (Table~\ref{tab:ds_stats}). Median token counts are 24 for questions and 130 for answers, aligning with the complexity of the curated WixQA datasets while providing greater quantity.
\end{itemize}

\subsection{Data Availability}
\label{sec:data_availability}
The WixQA datasets and KB corpus are publicly available as a Hugging Face Datasets \footnote{https://huggingface.co/datasets/Wix/WixQA} to encourage enterprise RAG research.

\setlength{\extrarowheight}{8pt}
\begin{table*}[htbp]
  \centering
  \setlength{\tabcolsep}{8pt} 
  \small
  \begin{tabular}{llcccccc} 
    \toprule
    \multicolumn{2}{c}{\textbf{Parameters}} & \multicolumn{6}{c}{\textbf{Metrics}} \\ 
    \cmidrule(lr){1-2} \cmidrule(lr){3-8} 
    \makecell{Retrieval\\Model} & \makecell{Generation\\Model} &
       \makecell{F1} &
       \makecell{BLEU} &
       \makecell{ROUGE-1} &
       \makecell{ROUGE-2} &
       \makecell{Context\\Recall} &
       \makecell{Factuality} \\ 
    \midrule
    bm25 & claude 3.7       & 0.37       & 0.09       & 0.31       & 0.12       & 0.73       & 0.80 \\ 
    bm25 & gemini 2.0 flash  & 0.39       & 0.12       & 0.32       & 0.15       & 0.72       & 0.72 \\ 
    bm25 & gpt 4o           & 0.36       & 0.08       & 0.29       & 0.12       & 0.73       & 0.83 \\ 
    bm25 & gpt 4o mini      & 0.37       & 0.08       & 0.30       & 0.11       & 0.72       & 0.76 \\ 
    e5   & claude 3.7       & 0.39       & 0.10       & 0.32       & 0.13       & \textbf{0.81}       & 0.82 \\ 
    e5   & gemini 2.0 flash  & \textbf{0.43} & \textbf{0.14} & \textbf{0.35} & \textbf{0.17} & \textbf{0.81}       & 0.76 \\ 
    e5   & gpt 4o           & 0.37       & 0.08       & 0.30       & 0.12       & \textbf{0.81}       & \textbf{0.85} \\ 
    e5   & gpt 4o mini      & 0.39       & 0.09       & 0.31       & 0.12       & \textbf{0.81}       & 0.79 \\ 
    \bottomrule
  \end{tabular}
  \caption{Performance on the \textbf{ExpertWritten} dataset}
  \label{tab:native}
\end{table*}

\begin{table*}[htbp]
  \centering
  \setlength{\tabcolsep}{8pt}
  \small
  \begin{tabular}{llcccccc} 
    \toprule
    \multicolumn{2}{c}{\textbf{Parameters}} & \multicolumn{6}{c}{\textbf{Metrics}} \\ 
    \cmidrule(lr){1-2} \cmidrule(lr){3-8} 
    \makecell{Retrieval\\Model} & \makecell{Generation\\Model} &
       \makecell{F1} &
       \makecell{BLEU} &
       \makecell{ROUGE-1} &
       \makecell{ROUGE-2} &
       \makecell{Context\\Recall} &
       \makecell{Factuality} \\ 
    \midrule
    bm25 & claude 3.7       & 0.21       & 0.03       & 0.21       & 0.08       & 0.55       & 0.74 \\ 
    bm25 & gemini 2.0 flash  & 0.28       & 0.04       & 0.26       & 0.11       & 0.55       & 0.63 \\ 
    bm25 & gpt 4o           & 0.22       & 0.03       & 0.22       & 0.08       & 0.55       & 0.76 \\ 
    bm25 & gpt 4o mini      & 0.23       & 0.03       & 0.21       & 0.08       & 0.55       & 0.74 \\ 
    e5   & claude 3.7       & 0.23       & 0.04       & 0.21       & 0.08       & \textbf{0.67} & \textbf{0.77} \\ 
    e5   & gemini 2.0 flash  & \textbf{0.30} & \textbf{0.05} & \textbf{0.28} & \textbf{0.12} & \textbf{0.67} & 0.66 \\ 
    e5   & gpt 4o           & 0.24       & 0.03       & 0.23       & 0.09       & \textbf{0.67} & \textbf{0.77} \\ 
    e5   & gpt 4o mini      & 0.24       & 0.04       & 0.23       & 0.09       & \textbf{0.67} & 0.75 \\ 
    \bottomrule
  \end{tabular}
  \caption{Performance on the \textbf{Simulated} dataset} 
  \label{tab:simulated}
\end{table*}

\begin{table*}[htbp]
  \centering
  \setlength{\tabcolsep}{8pt}
  \small
  \begin{tabular}{llcccccc} 
    \toprule
    \multicolumn{2}{c}{\textbf{Parameters}} & \multicolumn{6}{c}{\textbf{Metrics}} \\ 
    \cmidrule(lr){1-2} \cmidrule(lr){3-8} 
    \makecell{Retrieval\\Model} & \makecell{Generation\\Model} &
       \makecell{F1} &
       \makecell{BLEU} &
       \makecell{ROUGE-1} &
       \makecell{ROUGE-2} &
       \makecell{Context\\Recall} &
       \makecell{Factuality} \\ 
    \midrule
    bm25 & claude 3.7       & 0.42       & 0.18       & 0.38       & 0.22       & 0.95       & 0.84 \\ 
    bm25 & gemini 2.0 flash  & \textbf{0.59} & \textbf{0.33} & \textbf{0.53} & \textbf{0.37} & 0.96       & 0.85 \\ 
    bm25 & gpt 4o           & 0.47       & 0.20       & 0.42       & 0.24       & 0.95       & 0.86 \\ 
    bm25 & gpt 4o mini      & 0.43       & 0.17       & 0.38       & 0.19       & 0.95       & 0.84 \\ 
    e5   & claude 3.7       & 0.43       & 0.18       & 0.38       & 0.22       & \textbf{0.97} & 0.83 \\ 
    e5   & gemini 2.0 flash  & 0.58       & \textbf{0.33} & \textbf{0.53} & 0.36       & \textbf{0.97} & 0.84 \\ 
    e5   & gpt 4o           & 0.48       & 0.21       & 0.44       & 0.26       & 0.96       & \textbf{0.87} \\ 
    e5   & gpt 4o mini      & 0.44       & 0.18       & 0.39       & 0.20       & \textbf{0.97} & 0.86 \\ 
    \bottomrule
  \end{tabular}
  \caption{Performance on the \textbf{Synthetic} dataset}
  \label{tab:synthetic}
\end{table*}

\section{Experiments}
\label{sec:baseline_experiments}
In this section, we establish a benchmark baseline for retrieval-augmented generation (RAG) in enterprise support settings. Our evaluation spans the three distinct datasets (ExpertWritten, Simulated, and Synthetic) and leverages a combination of classical and dense retrieval methods alongside multiple state-of-the-art generation models. The goal of this benchmark creation is to provide a robust foundation for future research in procedural, multi-document QA. The pipeline was implemented using the FlashRAG tool~\cite{flashrag}, and executed in a single run, with specific configuration parameters detailed in Appendix~\ref{sec:flashrag_parameters}.

\subsection{Retrieval Methods}
We employed two standard retrieval approaches: keyword-based BM25 \cite{bm25} and semantic-based E5 dense retrieval (specifically, \texttt{e5-large-v2}, 335M parameters) \cite{wang2024textembeddingsweaklysupervisedcontrastive}. For retrieval, the top $k=5$ documents were selected. 

\subsection{Generation Models}
Our benchmark tests generation components that synthesize retrieved context into coherent, detailed procedural answers. We evaluate several state-of-the-art models—Claude 3.7, Gemini 2.0 Flash, GPT-4o, and GPT-4o Mini—running each with its default configuration to ensure a fair comparison across diverse architectures.

These models vary in language modeling and reasoning, affecting fluency, coherence, and multi-step explanation ability. This comparison reveals strengths and trade-offs for generating procedural responses aligned with real-world queries.

\subsection{Evaluation Metrics}
Due to the multi-faceted and procedural nature of the answers, we rely on a diverse set of evaluation metrics:
\begin{itemize}[noitemsep, topsep=2pt]
    \item \textbf{F1:} Token-level F1 score computed between the generated and gold answers.
    \item \textbf{BLEU:} An n-gram overlap metric that quantifies the similarity between the generated and the gold answers~\cite{papineni2002bleu}.
    \item \textbf{ROUGE-1 and ROUGE-2:} Metrics that capture unigram and bigram overlaps, respectively, to assess answer adequacy~\cite{lin2004rouge}.
    \item \textbf{Factuality:} An LLM-based judge metric \cite{llmasajudge} evaluating the factual alignment between the generated answer and the ground truth answer (see Appendix~\ref{app:factuality_prompt}). The LLM judge receives the original query, the generated answer, and the ground truth answer, and produces a score on a 0-1 scale based on how accurately the generated answer reflects the essential factual information present in the ground truth. This assesses whether the generator accurately utilized the provided context and avoided introducing factual errors or hallucinations. For our experiments, we utilized GPT-4o as the LLM judge for this metric.
    \item \textbf{Context Recall:} An LLM-based judge metric assessing the relationship between the retrieved context and the ground truth answer. To evaluate this, an LLM judge is provided with the user query, the retrieved context, and the ground truth answer. Following a specific instructional prompt (see Appendix~\ref{app:context_recall_prompt}), the LLM's task is to evaluate the extent to which the essential information required to formulate the ground truth answer is present within the retrieved context. It breaks down the ground truth answer into its core informational components and checks for their presence in the context, ignoring any additional, non-essential information within the context itself. Based on this analysis of information coverage, the LLM assigns a score on a 0-1 scale, reflecting the degree to which the retrieved context contains the necessary information to construct the ground truth answer. For our experiments, we utilized GPT-4o as the LLM judge for this metric as well.
\end{itemize}

\subsection{Results and Benchmark Baseline}
Tables~\ref{tab:native}, \ref{tab:simulated}, and \ref{tab:synthetic} summarize the performance of the RAG pipeline on the ExpertWritten, Simulated, and Synthetic datasets, respectively. Our baseline results yield several key insights:
\begin{itemize}[noitemsep, topsep=2pt]
    \item \textbf{Dense Retrieval Boosts Recall for Complex Queries:} The E5 dense retriever consistently outperforms BM25 on Context Recall, particularly for the ExpertWritten and Simulated datasets requiring multi-article synthesis. This highlights the benefit of semantic matching for complex information needs.

    \item \textbf{Dataset Difficulty Varies:} Performance differs markedly across datasets, with Synthetic yielding the highest scores, followed by ExpertWritten, and then Simulated proving the most challenging. This suggests varying difficulty levels related to query authenticity, answer complexity, and grounding requirements (single vs. multi-article).

    \item \textbf{Generation Models Exhibit Trade-offs:} No single generator excels universally. Models show distinct performance profiles, with some favouring n-gram similarity (F1, BLEU, ROUGE) while others achieve higher Factuality scores, indicating clear trade-offs relevant to specific application goals.

    \item \textbf{Enterprise RAG Requires Further Advancement:} While viable, the baseline scores, especially on ExpertWritten and Simulated datasets, underscore the significant challenge of generating accurate procedural answers from enterprise knowledge and highlight the need for continued research using benchmarks like WixQA.
\end{itemize}

These baselines show our RAG pipeline's viability for complex enterprise queries and offer a strong benchmark for future retrieval and generation research.

\section{Conclusion and Future Work}
\label{sec:conclusion}
To advance enterprise Retrieval-Augmented Generation (RAG) systems, we introduced \textbf{WixQA}, a benchmark suite featuring three QA datasets (ExpertWritten, Simulated, and Synthetic) and a 6,221-article knowledge base. WixQA's enables assessment of domain-specific procedural QA, particularly tasks requiring multi-document synthesis and complex, long-form answers. Our comprehensive baseline experiments establish initial performance levels and, critically, highlight persistent challenges in enterprise RAG.

Future work will focus on scaling these datasets, introducing multi-hop retrieval tasks, and refining human evaluation protocols. We release WixQA and the associated knowledge base (publicly available on Hugging Face) to foster advancements in RAG systems for reliable, user-centric enterprise applications and to provide a strong baseline for future research.

\section*{Acknowledgments}

We would like to extend our sincere gratitude to the dedicated efforts of our dataset labelers, including the Wix.com Customer Care Experts. Their meticulous attention to detail and unwavering commitment have been instrumental in refining the quality and accuracy of our datasets. We deeply appreciate the contributions of Serhii Shytikov, Kseniia Nalyvaiko, Kateryna Kharlamova, Anastasiia Bikovska, Inha Zatoka, Roy Shalish, and Ariel Yaakobi, whose expertise and hard work have significantly enhanced this research.

\bibliography{ref}

\begin{thebibliography}{22}
\providecommand{\natexlab}[1]{#1}

\bibitem[{Bolotova-Baranova et~al.(2023)Bolotova-Baranova, Blinov, Filippova, Scholer, and Sanderson}]{bolotova2023wikihowqa}
Valeriia Bolotova-Baranova, Vladislav Blinov, Sofya Filippova, Falk Scholer, and Mark Sanderson. 2023.
\newblock \href {https://doi.org/10.18653/v1/2023.acl-long.290} {{W}iki{H}ow{QA}: A comprehensive benchmark for multi-document non-factoid question answering}.
\newblock In \emph{Proceedings of the 61st Annual Meeting of the Association for Computational Linguistics (Volume 1: Long Papers)}, pages 5291--5314, Toronto, Canada. Association for Computational Linguistics.

\bibitem[{Campos et~al.(2020)Campos, Otegi, Soroa, Deriu, Cieliebak, and Agirre}]{doqa}
Jon~Ander Campos, Arantxa Otegi, Aitor Soroa, Jan Deriu, Mark Cieliebak, and Eneko Agirre. 2020.
\newblock \href {https://arxiv.org/abs/2005.01328} {Doqa -- accessing domain-specific faqs via conversational qa}.
\newblock \emph{Preprint}, arXiv:2005.01328.

\bibitem[{Castelli et~al.(2019)Castelli, Chakravarti, Dana, Ferritto, Florian, Franz, Garg, Khandelwal, McCarley, McCawley, Nasr, Pan, Pendus, Pitrelli, Pujar, Roukos, Sakrajda, Sil, Uceda-Sosa, Ward, and Zhang}]{techqa}
Vittorio Castelli, Rishav Chakravarti, Saswati Dana, Anthony Ferritto, Radu Florian, Martin Franz, Dinesh Garg, Dinesh Khandelwal, Scott McCarley, Mike McCawley, Mohamed Nasr, Lin Pan, Cezar Pendus, John Pitrelli, Saurabh Pujar, Salim Roukos, Andrzej Sakrajda, Avirup Sil, Rosario Uceda-Sosa, and 2 others. 2019.
\newblock \href {https://arxiv.org/abs/1911.02984} {The techqa dataset}.
\newblock \emph{Preprint}, arXiv:1911.02984.

\bibitem[{Chen et~al.(2024)Chen, Zhou, Hua, Xin, Chen, Li, Zhu, and Liang}]{fintextqa}
Jian Chen, Peilin Zhou, Yining Hua, Loh Xin, Kehui Chen, Ziyuan Li, Bing Zhu, and Junwei Liang. 2024.
\newblock \href {https://doi.org/10.18653/v1/2024.acl-long.328} {Fintextqa: A dataset for long-form financial question answering}.
\newblock In \emph{Proceedings of the 62nd Annual Meeting of the Association for Computational Linguistics (Volume 1: Long Papers)}, page 6025–6047. Association for Computational Linguistics.

\bibitem[{Chen et~al.(2022)Chen, Chen, Smiley, Shah, Borova, Langdon, Moussa, Beane, Huang, Routledge, and Wang}]{finqa}
Zhiyu Chen, Wenhu Chen, Charese Smiley, Sameena Shah, Iana Borova, Dylan Langdon, Reema Moussa, Matt Beane, Ting-Hao Huang, Bryan Routledge, and William~Yang Wang. 2022.
\newblock \href {https://arxiv.org/abs/2109.00122} {Finqa: A dataset of numerical reasoning over financial data}.
\newblock \emph{Preprint}, arXiv:2109.00122.

\bibitem[{Feng et~al.(2015)Feng, Xiang, Glass, Wang, and Zhou}]{feng2015insuranceqa}
Minwei Feng, Bing Xiang, Michael~R. Glass, Lidan Wang, and Bowen Zhou. 2015.
\newblock \href {https://arxiv.org/abs/1508.01585} {Applying deep learning to answer selection: A study and an open task}.
\newblock \emph{Preprint}, arXiv:1508.01585.

\bibitem[{Feng et~al.(2021)Feng, Patel, Wan, and Joshi}]{multidoc2dial}
Song Feng, Siva~Sankalp Patel, Hui Wan, and Sachindra Joshi. 2021.
\newblock \href {https://doi.org/10.18653/v1/2021.emnlp-main.498} {{M}ulti{D}oc2{D}ial: Modeling dialogues grounded in multiple documents}.
\newblock In \emph{Proceedings of the 2021 Conference on Empirical Methods in Natural Language Processing}, pages 6162--6176, Online and Punta Cana, Dominican Republic. Association for Computational Linguistics.

\bibitem[{Feng et~al.(2020)Feng, Wan, Gunasekara, Patel, Joshi, and Lastras}]{doc2dial}
Song Feng, Hui Wan, Chulaka Gunasekara, Siva Patel, Sachindra Joshi, and Luis Lastras. 2020.
\newblock \href {https://doi.org/10.18653/v1/2020.emnlp-main.652} {doc2dial: A goal-oriented document-grounded dialogue dataset}.
\newblock In \emph{Proceedings of the 2020 Conference on Empirical Methods in Natural Language Processing (EMNLP)}, pages 8118--8128, Online. Association for Computational Linguistics.

\bibitem[{Gupta et~al.(2019)Gupta, Kulkarni, Chanda, Rayasam, and Lipton}]{gupta2019amazonqa}
Mansi Gupta, Nitish Kulkarni, Raghuveer Chanda, Anirudha Rayasam, and Zachary~C Lipton. 2019.
\newblock \href {https://arxiv.org/abs/1908.04364} {Amazonqa: A review-based question answering task}.
\newblock \emph{Preprint}, arXiv:1908.04364.

\bibitem[{Jin et~al.(2024)Jin, Zhu, Yang, Zhang, and Dou}]{flashrag}
Jiajie Jin, Yutao Zhu, Xinyu Yang, Chenghao Zhang, and Zhicheng Dou. 2024.
\newblock \href {https://doi.org/10.48550/ARXIV.2405.13576} {Flashrag: {A} modular toolkit for efficient retrieval-augmented generation research}.
\newblock \emph{CoRR}, abs/2405.13576.

\bibitem[{Jin et~al.(2019)Jin, Dhingra, Liu, Cohen, and Lu}]{jin2019pubmedqa}
Qiao Jin, Bhuwan Dhingra, Zhengping Liu, William~W. Cohen, and Xinghua Lu. 2019.
\newblock \href {https://arxiv.org/abs/1909.06146} {Pubmedqa: A dataset for biomedical research question answering}.
\newblock \emph{Preprint}, arXiv:1909.06146.

\bibitem[{Joshi et~al.(2017)Joshi, Choi, Weld, and Zettlemoyer}]{triviaqa}
Mandar Joshi, Eunsol Choi, Daniel~S. Weld, and Luke Zettlemoyer. 2017.
\newblock \href {https://arxiv.org/abs/1705.03551} {Triviaqa: A large scale distantly supervised challenge dataset for reading comprehension}.
\newblock \emph{Preprint}, arXiv:1705.03551.

\bibitem[{Kwiatkowski et~al.(2019)Kwiatkowski, Palomaki, Redfield, Collins, Parikh, Alberti, Epstein, Polosukhin, Devlin, Lee, Toutanova, Jones, Kelcey, Chang, Dai, Uszkoreit, Le, and Petrov}]{nq}
Tom Kwiatkowski, Jennimaria Palomaki, Olivia Redfield, Michael Collins, Ankur Parikh, Chris Alberti, Danielle Epstein, Illia Polosukhin, Jacob Devlin, Kenton Lee, Kristina Toutanova, Llion Jones, Matthew Kelcey, Ming-Wei Chang, Andrew~M. Dai, Jakob Uszkoreit, Quoc Le, and Slav Petrov. 2019.
\newblock \href {https://doi.org/10.1162/tacl_a_00276} {Natural questions: A benchmark for question answering research}.
\newblock \emph{Transactions of the Association for Computational Linguistics}, 7:452--466.

\bibitem[{Lewis et~al.(2021)Lewis, Perez, Piktus, Petroni, Karpukhin, Goyal, Küttler, Lewis, tau Yih, Rocktäschel, Riedel, and Kiela}]{lewis2020rag}
Patrick Lewis, Ethan Perez, Aleksandra Piktus, Fabio Petroni, Vladimir Karpukhin, Naman Goyal, Heinrich Küttler, Mike Lewis, Wen tau Yih, Tim Rocktäschel, Sebastian Riedel, and Douwe Kiela. 2021.
\newblock \href {https://arxiv.org/abs/2005.11401} {Retrieval-augmented generation for knowledge-intensive nlp tasks}.
\newblock \emph{Preprint}, arXiv:2005.11401.

\bibitem[{Lin(2004)}]{lin2004rouge}
Chin-Yew Lin. 2004.
\newblock \href {https://aclanthology.org/W04-1013/} {{ROUGE}: A package for automatic evaluation of summaries}.
\newblock In \emph{Text Summarization Branches Out}, pages 74--81, Barcelona, Spain. Association for Computational Linguistics.

\bibitem[{Papineni et~al.(2002)Papineni, Roukos, Ward, and Zhu}]{papineni2002bleu}
Kishore Papineni, Salim Roukos, Todd Ward, and Wei-Jing Zhu. 2002.
\newblock \href {https://doi.org/10.3115/1073083.1073135} {{B}leu: a method for automatic evaluation of machine translation}.
\newblock In \emph{Proceedings of the 40th Annual Meeting of the Association for Computational Linguistics}, pages 311--318, Philadelphia, Pennsylvania, USA. Association for Computational Linguistics.

\bibitem[{Rajpurkar et~al.(2016)Rajpurkar, Zhang, Lopyrev, and Liang}]{squad}
Pranav Rajpurkar, Jian Zhang, Konstantin Lopyrev, and Percy Liang. 2016.
\newblock \href {https://arxiv.org/abs/1606.05250} {Squad: 100,000+ questions for machine comprehension of text}.
\newblock \emph{Preprint}, arXiv:1606.05250.

\bibitem[{Robertson and Zaragoza(2009)}]{bm25}
Stephen Robertson and Hugo Zaragoza. 2009.
\newblock \href {https://doi.org/10.1561/1500000019} {The probabilistic relevance framework: Bm25 and beyond}.
\newblock \emph{Foundations and Trends in Information Retrieval}, 3:333--389.

\bibitem[{Tan et~al.(2025)Tan, Zhuang, Montgomery, Tang, Cuadron, Wang, Popa, and Stoica}]{llmasajudge}
Sijun Tan, Siyuan Zhuang, Kyle Montgomery, William~Y. Tang, Alejandro Cuadron, Chenguang Wang, Raluca~Ada Popa, and Ion Stoica. 2025.
\newblock \href {https://arxiv.org/abs/2410.12784} {Judgebench: A benchmark for evaluating llm-based judges}.
\newblock \emph{Preprint}, arXiv:2410.12784.

\bibitem[{Wang et~al.(2024{\natexlab{a}})Wang, Yang, Huang, Jiao, Yang, Jiang, Majumder, and Wei}]{wang2024textembeddingsweaklysupervisedcontrastive}
Liang Wang, Nan Yang, Xiaolong Huang, Binxing Jiao, Linjun Yang, Daxin Jiang, Rangan Majumder, and Furu Wei. 2024{\natexlab{a}}.
\newblock \href {https://arxiv.org/abs/2212.03533} {Text embeddings by weakly-supervised contrastive pre-training}.
\newblock \emph{Preprint}, arXiv:2212.03533.

\bibitem[{Wang et~al.(2024{\natexlab{b}})Wang, Liu, Song, Cheng, Fu, Guo, Fang, Zhu, and Dou}]{domainrag}
Shuting Wang, Jiongnan Liu, Shiren Song, Jiehan Cheng, Yuqi Fu, Peidong Guo, Kun Fang, Yutao Zhu, and Zhicheng Dou. 2024{\natexlab{b}}.
\newblock \href {https://arxiv.org/abs/2406.05654} {Domainrag: A chinese benchmark for evaluating domain-specific retrieval-augmented generation}.
\newblock \emph{Preprint}, arXiv:2406.05654.

\bibitem[{Ye et~al.(2024)Ye, Lei, Yin, Chen, Zhou, and He}]{convrag}
Linhao Ye, Zhikai Lei, Jianghao Yin, Qin Chen, Jie Zhou, and Liang He. 2024.
\newblock \href {https://arxiv.org/abs/2403.18243} {Boosting conversational question answering with fine-grained retrieval-augmentation and self-check}.
\newblock \emph{Preprint}, arXiv:2403.18243.

\end{thebibliography}

\clearpage
\appendix

\section{Instructions for Expert Answer Authoring}
\label{app:expert_authoring_guidelines}
When authoring answers based on Knowledge Base (KB) articles, you were to adhere to the following instructions:

\begin{itemize}[noitemsep, topsep=2pt]
    \item \textbf{Provide Relevant KB Article URLs:} Identify and provide the URLs of all relevant KB articles that, in combination, fully answer the user's question. For data collection, each URL was to be listed in a separate column.
    \item \textbf{Write a Full and Relevant Answer:} Compose a full answer. A full answer provides only the relevant information, and all the relevant information, necessary for users to resolve their issues. Information not pertinent to the solution should be omitted.
    \item \textbf{Adhere Strictly to KB Content:} Copy the answer from the identified KB articles as much as possible.
    \item \textbf{Exclusive Reliance on KB:} Do not use any information from your previous experience that does not rely on the provided KB.
    \item \textbf{User-Centric Phrasing and Linking:} Phrase your answer as you would to a real user. When needed and supported by the KB material, add links within the answer (e.g., to other KB articles, specific Wix resources, or pertinent external sources).
\end{itemize}

\section{QA Examples}
\label{app:QA_examples}

\subsection{WixQA-ExpertWritten Example}
\label{app:ExpertWritten_example}
Below is a representative example of a QA in the WixQA-ExpertWritten dataset:

\begin{quote}
\small
\setlength{\parindent}{0pt}

QUESTION 

I cant access my old email address to change my password. I have a recovery phone number added to my Wix account.

ANSWER

If you've previously added a Recovery Phone Number to your Wix account, you can get a one-time SMS code sent to your mobile phone to recover your Wix account.
  
To recover your account:

1. Go to the Wix Log In page. (https://users.wix.com/signin)
  
2. Click Forgot Email?.
  
3. Select Can't access my login email.
  
4. Click Next.
  
5. Enter the email address associated with your Wix account. 
  
6. Click Next.
  
7. Select Send me a code to my phone number ending with XXXX  and click Next.

8. Enter the one-time verification code sent to your phone under Enter code here.
  
9. Click Verify.
  
10. Enter a new password and retype it.
  
11. Click Save New Password. 
  
After you recover your account, you should change the email address of your account (https://support.wix.com/en/article/changing-the-email-address-of-your-wix-account) to one that you are able to access.
  
To change your email address:
  
1. Go to Account Settings in your Wix account.

2. Click the Account email field.
  
3. Enter your Wix account password in the pop-up.
  
4. Enter your new account email address.
  
5. Retype your new email account email address to confirm it.
  
6. Click Next.
  
7. Check your email account for an email from Wix.
  
8. Copy the 6-digit confirmation code.
  
9. Go back to Account Settings in your Wix account.
  
10. Paste the code into the pop-up.
  
11. Click Submit.

ARTICLES

"391586...",
"e93664..."

\end{quote}

\subsection{WixQA-Simulated Example}
\label{app:Simulated_example}

Below is a representative example of a QA in the WixQA-Simulated dataset:

\begin{quote}
\small
\setlength{\parindent}{0pt}

QUESTION 

How do I link a button to the users member page without linking to a dataset?

ANSWER

To link a button to a members page without using a dataset, add a button in the Wix Editor, click the Link icon, select Page, choose the members page from the list, and click Done.

ARTICLES

"926ba4...",
"fc77f8..."
\end{quote}

\subsection{WixQA-Synthetic Example}
\label{app:Synthetic_example}

Below is a representative example of a QA in the WixQA-Synthetic: dataset:

\begin{quote}
\small
\setlength{\parindent}{0pt}

QUESTION 

Why is my blog post showing the full content instead of just a preview when connected to pricing plans?

ANSWER

This is a known problem where blog posts connected to pricing plans show the full post instead of the preview. As a temporary workaround, you can remove the preview entirely by following these steps:

1. Enter Post composer for Post

2. Click monetize

3. Move preview to the top to remove the entire preview

4. Republish post

5. Refresh live site

ARTICLES

"3835e0..."

\end{quote}

\section{LLM Prompt for Factuality Evaluation}
\label{app:factuality_prompt}
Below is the prompt used to evaluate the factual consistency (i.e., actuality) between the generated answer and the golden answer:

\begin{quote}
\small
\setlength{\parindent}{0pt}

ROLE

You are a Factual Alignment Expert. Your job is to evaluate how well an AI response includes the essential information from a ground truth answer (GT answer) according to a given user query.

Note that the Ground Truth (GT Answer), is the "Correct" answer generated by an expert, and was created to evaluate the model, and is NOT part of the AI response or the context.

TASK DESCRIPTION

You will be presented with three elements: a question, a GT answer, and an AI response. Determine how well the AI response includes the essential information from the GT answer that helps to solve the user's query.
In case of  any additional or extra information present in the AI response, only make sure it's not preventing the user from solving his query.

EVALUATION CRITERIA

5: Complete Match - All essential information from GT answer appears in AI response, providing complete solution to the query

4: Strong Match - Most essential information is present, with only minor details missing that don't impact the solution significantly

3: Partial Match - Core information is present but missing some important details that would help better solve the query

2: Limited Match - Only basic or partial information present, missing several essential elements needed for the solution

1: Poor Match - Missing most essential information or contains incorrect information that could mislead the user

INSTRUCTIONS

1. Read the question carefully and analyze the ground truth answer to identify all key information elements that help solve the query

2. Compare the AI response (candidate answer) against the ground truth, focusing on presence of important information

3. Evaluate the completeness and accuracy of the information transfer

4. Assign a rating (0-1) based on how well important information is preserved

5. Provide a brief explanation focusing on factuality
\end{quote}

\section{LLM Prompt for Context Recall Evaluation}
\label{app:context_recall_prompt}
Below is the prompt used to evaluate the context recall between the retrieved context and the golden answer:

\begin{quote}
\small
\setlength{\parindent}{0pt}
ROLE 
You are a Context Evaluation Expert. Your job is to assess how well a retrieved context contains the essential information present in a ground truth answer (GT answer).

TASK DESCRIPTION
You will be presented with three pieces of information: a user query, its ground truth answer , and a retrieved context (that will be used to create an AI response from). Determine how well the essential information from the GT answer appears in the retrieved context. Additional information in the retrieved context should not affect the scoring.

EVALUATION CRITERIA

5: Complete Match - All essential information from the GT answer is present in the retrieved context. The context fully enables answering the user's question.
4: Strong Match - All essential information is present, but some minor details are missing. The context still effectively answers the user's question.
3: Partial Match - Most essential information is present, but some important details are missing. The context partially answers the user's question.
2: Weak Match - Only basic or limited essential information is present. The context provides insufficient information to properly answer the user's question.
1: No Match - Essential information is missing or incorrect. The context cannot be used to answer the user's question.

INSTRUCTIONS

1. Read the question to understand the idea of what the user asks for
2. Break down the GT answer into essential information (key facts, main concepts, direct answers).
3. Check if these information pieces appear in the retrieved context
4. Focus only on finding the ground truth information in the context - ignore any additional or extra information present in the retrieved context
5. Assign a rating (0-1) based on information coverage and relevance

\end{quote}

\section{LLM Prompt for QA pairs extraction from Feature Request articles}
\label{app:qa_extraction_feature_request}
Below is a prompt for GPT-4o we used to extract question--answer pairs for Feature Request type of articles:

\begin{quote}
\small
\setlength{\parindent}{0pt}

The kb article provided below contains information about a feature or an action that may not be implemented yet. Find the not implemented feature from the article, ask a plausible user question if the feature is supported, and answer it using the article.

Ensuring the following:

1. Use the exact wording from the article for the answer whenever possible.

2. Avoid any hallucinations at the end, such as "For more information..."

3. If there are relevant step-by-step instructions, include them in the answer.

4. Do not skip the information from the step-by-step instructions.

5. If there is a relevant workaround or tip, include it.

6. Copy relevant links directly from the context as they are.

7. Do not include phrases suggesting to contact support unless absolutely necessary.

8. Ensure that the image names are not used as usual text.

9. Do not include: "We are always working to update and improve our products, and your feedback is hugely appreciated".

10. Do not include announcement like "We are excited to announce".

Provide the output as JSON with "question" and "answer" fields. Format the "answer" field value as a markdown.

\end{quote}

\section{FlashRAG Configuration Parameters} \label{sec:flashrag_parameters}

This section lists the core FlashRAG configuration parameters used, corresponding to the framework's default settings. Parameters related to file paths and environment specifics have been omitted. Note that specific hyperparameters tuned during our experiments (such as the generation model) are not detailed here.

\begin{itemize}
    \item generator\_batch\_size: 25
    \item generator\_max\_input\_len: 50,000
    \item retrieval\_topk: 5
    \item retrieval\_query\_max\_length: 10,000
    \item retrieval\_batch\_size: 1024
    \item max\_tokens: 1024
    \item temperature: 0
\end{itemize}

\end{document}